\documentclass[letterpaper, 10 pt, conference]{ieeeconf}  % Comment this line out if you need a4paper

\IEEEoverridecommandlockouts                              % This command is only needed if 
\usepackage{graphicx} % for pdf, bitmapped graphics files
\usepackage{amsmath} % assumes amsmath package installed
\usepackage{amssymb}  % assumes amsmath package installed
\usepackage{bm} % for bold letters/symbols in math mode
\usepackage{dsfont} % for the R symbol (real numbers)
\usepackage{subfig}	% for subfigures
\usepackage{todonotes}
\usepackage{hyperref}

\title{\LARGE \bf
Comparison of Distal Teacher Learning with Numerical and Analytical Methods to Solve Inverse Kinematics for Rigid-Body Mechanisms
}

\author{Tim von Oehsen$^{1}$, Alexander Fabisch$^{1,2}$, Shivesh Kumar$^{2}$ and Frank Kirchner$^{1,2}$% <-this % stops a space
\thanks{*This work was supported by the German Federal Ministry for Economic Affairs and Energy
(BMWi, grant no. 50RA1701) and the German Federal Ministry of Education and Research (BMBF, grant no. 01IW18003).}% <-this % stops a space
\thanks{$^{1}$Robotics Research Group, University of Bremen, Robert-Hooke-Str. 1, D-28359 Bremen, Germany
        {\tt\small tim\_von@uni-bremen.de}}%
\thanks{$^{2}$Robotics Innovation Center, German Research Center for Artificial Intelligence (DFKI GmbH), Robert-Hooke-Str. 1, D-28359 Bremen, Germany
        {\tt\small \{shivesh.kumar,alexander.fabisch\}@dfki.de}}%
}

\begin{document}

\maketitle
\thispagestyle{empty}
\pagestyle{empty}

%%%%%%%%%%%%%%%%%%%%%%%%%%%%%%%%%%%%%%%%%%%%%%%%%%%%%%%%%%%%%%%%%%%%%%%%%%%%%%%%
\begin{abstract}

Several publications are concerned with learning inverse kinematics, however,
their evaluation is often limited and none of the proposed methods is of
practical relevance for rigid-body kinematics with a known forward model.
We argue that for rigid-body kinematics one of the first proposed
machine learning (ML) solutions to inverse kinematics -- distal teaching (DT) -- is actually
good enough when combined with differentiable programming libraries and we
provide an extensive evaluation and comparison to analytical and numerical solutions.
In particular, we analyze solve rate, accuracy, sample efficiency and scalability. Further, we
study how DT handles joint limits, singularities, unreachable
poses, trajectories and provide a comparison of execution times.
The three approaches are evaluated on three different rigid body mechanisms with varying complexity.
With enough training data and relaxed precision requirements, DT
has a better solve rate and is faster than state-of-the-art numerical solvers for a 15-DoF
mechanism.
DT is not affected by singularities while numerical solutions
are vulnerable to them.
In all other cases numerical solutions are usually better. Analytical solutions
outperform the other approaches by far if they are available.

\end{abstract}

%%%%%%%%%%%%%%%%%%%%%%%%%%%%%%%%%%%%%%%%%%%%%%%%%%%%%%%%%%%%%%%%%%%%%%%%%%%%%%%%
\section{Introduction}

Human beings are extraordinary in performing movements. They can describe and perform these 
movements in their physical space without thinking about how to control every individual muscle.
Robots are systems built from various links and actuators capable of performing movements.
Roboticists often control individual actuators directly to interact with the world.
To describe a movement in physical space or in task space we need a relationship to convert the 
task space configuration into the joint space configuration. This is called the inverse 
kinematics problem (IKP) and is a central problem in robot control. It is about
calculating joint angles $\bm{\theta} \in \mathds{R}^n$ of a kinematic chain
to reach a certain end effector pose $\bm{x} \in \mathds{R}^m$. $n$ and $m$
denote the degrees of freedom (DoF) of the joint space and work space respectively. 
IKP is usually easy to formulate but difficult to solve for robots with serial architecture
as it typically results in a set of non-linear algebraic equations. 
Often there are multiple solutions to this problem and their complexity is highly dependent on the geometry of the robot.
One of the early efforts to solve the IKP was by Pieper~\cite{pieper1968kinematics}.
He gave closed-form solutions for 6-DoF manipulators with three revolute or prismatic joints whose axes intersect at a point.
Similarly, Paul~\cite{1982_Paul_book} developed methods to compute analytical solutions for robots with relatively simple geometry 
that have many zero distances and parallel or perpendicular joint axes. 
For the IKP of a 6R robot with arbitrary geometry, Raghavan and Roth~\cite{raghavan1993inverse} 
derived a 16 degree univariate polynomial which provides all possible solutions. 
When an explicit solution to the IKP is not available or difficult to obtain in real-time, 
iterative numerical techniques based on Newton-Raphson or inverse Jacobian-based methods can be utilized.

Despite the existence of analytical and numerical solutions to solve this task, there is
active research in the field of machine learning to solve the IKP.
But why should we learn inverse kinematics?
Learning is beneficial when the kinematic model is completely or
partially unknown. In this case, traditional analytical or numerical solutions
are less accurate or not even available because they depend on a model.
Machine learning methods enable us to learn IK even without a known kinematic
model though. This is why most work focuses on learning IK from scratch.
However, to this date, serial kinematic chains with rigid-body geometries are still
the most common type of robots employed in real-world applications.
For these specific robots the forward kinematics are known very accurately.
Despite rigid-body mechanisms being a very common type, there have not been made
extensive evaluations in this regard yet and it is not clear whether machine learning
also has potential to be a valid alternative to analytical or numerical solutions.
Hence, in this work the main focus lies on the comparison of a machine learning solution
to an analytical and a numerical solution
under known forward kinematics (FK), that is,
$\bm{x} = \bm{f}(\bm{\theta})$.

Additionally, the assumed perfect knowledge of the FK makes it possible to evaluate
a machine learning approach to inverse kinematics in detail.
Gained insights can also (at least partially) be transferred to
more difficult variations of the problem like unknown forward kinematics,
flexible joints or links, torque-based gravity compensation or inverse dynamics.
These problems share some characteristics with the IKP
for known forward kinematics from the machine learning perspective:
(1) accurate model parameters are not available,
(2) training data has to cover the whole workspace,
(3) limits (e.g. joint limits) have to be respected,
and (4) computational efficiency is required for robot control.

\section{Related Work}

In general, the IKP has multiple solutions, even if the robot is \textit{properly actuated}, i.~e., $m=n$ holds.
The main problem to overcome when learning IK is the nonconvexity of
the solution set which prevents the application of simple direct regression techniques \cite{Jordan.1992} 
because they lead to an averaging over the nonconvex solution space.
A lot of machine learning approaches to solve inverse kinematics have
been published.
\textit{Locally Weighted Projection Regression} \cite{DSouza.29.10.2001}
works with respect to the differentiated forward kinematics, that is
$\bm{\dot{x}}=\bm{J}(\bm{\theta})  \bm{\dot{\theta}}$.
The matrix $\bm{J} = \frac{d\bm{f}}{d\bm{\theta}}$ is the Jacobian of the manipulator.
By doing so, the problem can be transformed into a locally convex one.
In \cite{Ardizzone.14.08.2018}, \textit{Invertible Neural Networks}
are proposed. They can be utilized to solve inverse kinematics by just
learning the simpler forward kinematics which leads to unique solutions.
To resolve the redundancy of IK solutions, latent variables
containing the necessary information are introduced.
The method \textit{Goal Babbling} \cite{Rolf.2010, Rolf.2014} aims to learn an IK model
by employing goal-directed movements along paths to a pose.
Removing redundant configurations during sampling resolves the nonconvexity problem.
B\'{o}csi et al. \cite{Bocsi.25.09.2011} learned the joint probability
density $p(\bm{x},\bm{\theta})$ via \textit{Structured Output Learning}.
There have also been ideas to apply deep reinforcement learning techniques
\cite{Phaniteja.05.12.2017}.
While the attempts have been quite diverse, none of these approaches has been extensively
examined and it is not clear whether there is potential to use them for
rigid-body mechanisms.
In particular, several of the aspects mentioned above have not been explored on
systems with varying complexity and no comparisons to analytical and numerical
methods were done.

\section{Investigated Solutions to the IKP}

%In the following sections we introduce the investigated approaches to %solve the IKP for rigid-body mechanisms.

\subsection{Analytical Solution}
It is important to note that analytical solutions are robot dependent and need to be derived individually for every mechanism.
They can only be obtained if certain conditions are met. For redundant robots ($n > m$) there are no general analytical solutions.
There are still conditions regarding the geometrical structure of the mechanism such that a general analytical solution exists
if the dimensionalities of the task and action space equal ($n = m$). An example for a 6-DoF arm is the intersection of the axes of three consecutive revolute joints in one point at the wrist.
Many manipulators in the industry use a spherical wrist which fulfills this requirement. For those mechanisms a geometrical closed-form solution is possible. The spherical wrist permits to split the IKP into a position and orientation equation that depend on three joints each and are solvable analytically.
%IKFast~\cite{2010_Diankov_IKFAST} analytically solves robot inverse kinematics equations and generates optimized C++ files from the general XML description of a robot. IKFast works with any number of joints arranged in a kinematic chain. For redundant kinematic chains, the user can set arbitrary values of a subset of the joints until $n = m$.

\subsection{Numerical Solution: TRAC-IK}
TRAC-IK \cite{Beeson.2015} is a state-of-the-art numerical solver. Like all numerical methods, it can be used for any serial kinematic chain with known geometrical parameters. The solver consists of a combination of two algorithms. The first one is based on the pseudo-inverse of the Jacobian matrix i.e. in case of redundant robots $(n > m)$, the solution is given by $\bm{\dot{\theta}}=\bm{J}^+(\bm{\theta})  \bm{\dot{x}}$ \cite{Buss.2004}. In case $n = m$, Jacobian inverse is sufficient.
% It refers to the first derivative of the FK function, namely $\bm{\dot{\theta}}=\bm{J}^+(\bm{\theta})  \bm{\dot{x}}$. 
Via an iterative procedure an IK solution can be approximated until the Cartesian error falls below a selected threshold. TRAC-IK further adds random restarts whenever the algorithm gets stuck in local minima.
The second part of the solver is based on \textit{Sequential Quadratic Programming} (SQP, \cite{Boggs.1995}). It is required to formulate the IKP as a two times differentiable optimization problem to apply SQP. In TRAC-IK, the objective function $\phi=\bm{p}_{err}^T  \bm{p}_{err}$ will be minimized, where $\bm{p}_{err} \in \mathds{R}^6$ contains Cartesian errors in position and orientation. Thus, the objective function is the sum of squared errors. In addition, joint limits are included as constraints.
The reason behind combining these two approaches is to benefit from the different advantages which are mainly the low execution time of the pseudo-inverse solver and the high solve rate of the SQP solver.

\subsection{Data-driven solution: Distal Teacher Learning}
\label{subsec:DTL}
Distal teacher learning \cite{Jordan.1992} -- we refer to it as \textit{distal teaching} (DT) in this paper -- is a data-driven method to solve inverse problems. It can be applied to the IKP as well and does not depend on a known FK model. However, the latter is usually known well for rigid-body mechanisms.
It can be used to simplify the training process.
The main idea of DT is to calculate the prediction error of the IK model in the Cartesian space by propagating predicted joint angle values through the FK model that acts as the distal teacher. This procedure is one way to handle the nonconvexity of the IKP. A direct calculation of prediction errors in the joint space (\textit{direct inverse modeling}) would lead to a very bad performance of the learned IK model because an averaging over the nonconvex solution space would occur. In DT, the goal is the identity mapping
\begin{equation}
f(g(\bm{x}^*)) = \bm{x}^*
\end{equation}
where $f$ is the FK function, $g$ the IK and $\bm{x}^*$ the desired pose. The difference of the predicted pose and the desired pose is the error signal used to optimize the IK model. If backpropagation is applied, this error signal needs to be passed backwards through the FK model without updating it and is then used to optimize the IK model parameters.
Under the assumption of a known FK function, we do not need to learn a FK model first and can directly optimize an IK model. For this purpose, we propose to integrate the FK function into the loss function of a feed-forward neural network with fully connected layers. The latter serves as the IK model. The implementation of this approach becomes very easy with modern differentiable programming libraries. In this case we use TensorFlow \cite{tensorflow2015whitepaper}. Due to its simplicity in case of a known FK function, DT is very well suited for learning the IK of rigid-body mechanisms.
In order to train the neural network, we need to define a suitable loss. In general, in this work the weighted sum
\begin{equation}
\label{eq:loss}
\varepsilon = w  \varepsilon_{pos} + (1-w)  \varepsilon_{ori}
\end{equation}
of the position error $\varepsilon_{pos}$ and the orientation error $\varepsilon_{ori}$ of the end effector is used. $w \in [0, 1]$ is a freely choosable hyperparameter, which we set to $0.75$ for our analysis. The position error is defined as the Euclidean distance of the desired pose and the predicted pose that results from propagating the predicted joint angles through the FK model. For the orientation error, we use the quaternion difference $\varepsilon_{ori} = 2 \, \textrm{acos}\left( | \hat{\bm{q}}^T  \bm{q}^* | \right)$ \cite{DuHuynh.2009} since we represent the orientations as quaternions ($\bm{q}$). Furthermore, it is important to note that we encode the predicted joint angles by using $\sin\bm{\theta}$ and $\cos\bm{\theta}$ which is a common technique to encode periodic variables \cite[pp. 105--110]{Bishop.2006}
and increases performance. Hence, the number of output nodes is doubled. Afterwards, joint angles can be decoded as $\bm{\theta} = \textrm{atan2}\left( \sin\bm{\theta}, \cos\bm{\theta} \right)$. %The \textrm{atan2}-function is necessary to cover all four quadrants.
In its basic form described above, DT does not take any measures regarding the joint limits \cite{Rolf.2010}. Hence, the learned IK outputs a lot of infeasible predictions. In order to handle joint limits adequately we propose to add a linear penalty term to the loss such that it becomes
\begin{equation}
\label{eq:limits_loss}
\varepsilon = w  \varepsilon_{pos} + (1-w)  \varepsilon_{ori} + \lambda  d_v.
\end{equation}
$\lambda$ is a hyperparameter that defines the strength of the penalty. The variable $d_v$ is the violation distance, i.~e., the sum of all distances to the next joint limit in infeasible joint regions.

\section{Methodology}

A main contribution of this paper is the rigorous evaluation and comparison of three different approaches to the IKP. We would like to encourage other researchers to use similar benchmarks. Our concept will be presented in this section. Our implementation can be found at \url{https://github.com/tvoeh/distal-teaching-ik}.

\subsection{Manipulators}
\begin{figure}[tb!]
\vskip -0.04in
\centering
\subfloat[6-DoF arm COMPI (source: \cite{Compi19}).]{
  \includegraphics[height=3cm,clip=true,trim=100 0 150 0]{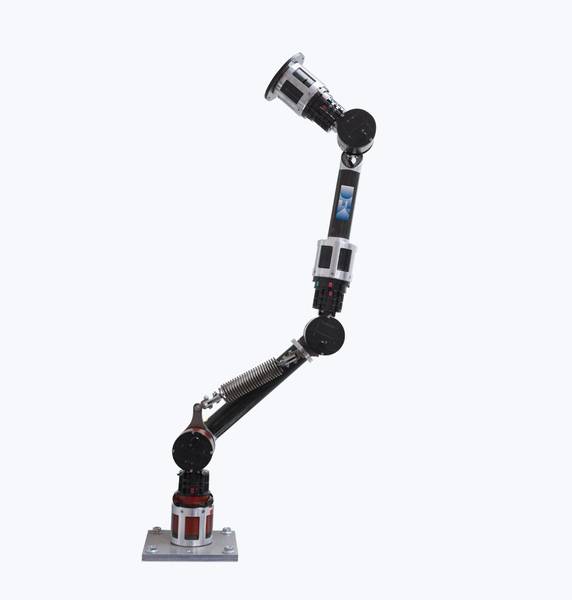}
	\label{fig:Compi}
}\hspace{0.5cm}
\subfloat[Model of the humanoid Atlas. A 15-DoF subchain (highlighted in red) from the left foot to the left hand is investigated.]{
  \includegraphics[height=3cm]{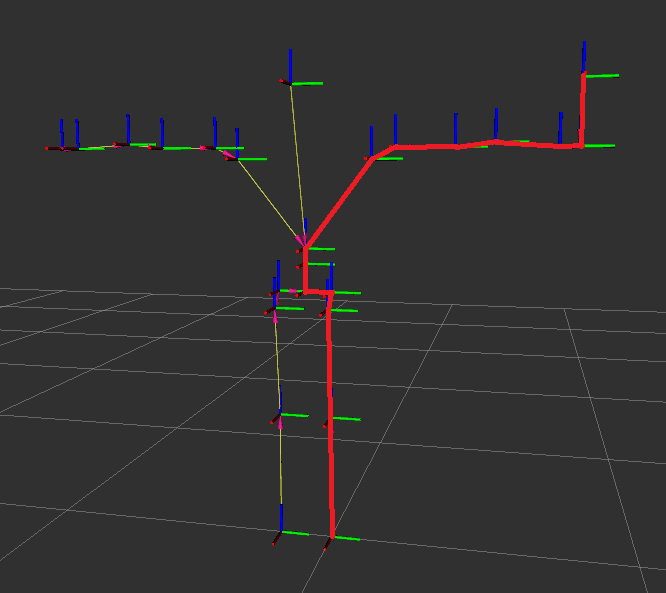}
	\label{fig:Atlas}
}
\caption{Evaluated mechanisms in 3D space.\label{fig:3D_mechanisms}}
\vskip -0.27in
\end{figure}

We use three manipulators with different complexities. All of them are serial kinematic chains with rigid links and revolute joints only. The first mechanism is a simple 3-DoF arm with parallel joint axes which operates in a plane. Its total length is \mbox{1.3 m}. The other two mechanisms work in 3D space and are shown in Fig. \ref{fig:3D_mechanisms}. One of them is the 6-DoF arm COMPI \cite{Bargsten2015} that possesses a spherical wrist and has a total length of \mbox{1.12 m}. The last mechanism is a 15-DoF subchain of the humanoid Atlas of Boston Dynamics. We investigate the serial chain from the left foot to the left hand where the foot is treated as the base. We further added a tool to the left hand which extends the total length of the chain to \mbox{2.31 m}.

\subsection{Experiment Design}
We conduct various experiments to compare DT to the analytical solution and the numerical solver TRAC-IK to get insights on when it is useful in case of a known FK function. We sample training and test data from a uniform distribution in joint space unless otherwise stated.
\paragraph{Sample efficiency} We investigate the effect of number of training samples on position and orientation errors.
\paragraph{Joint limits} We investigate handling of joint limits and use the loss function given in Eq. \ref{eq:limits_loss} for DT. Note that in all other experiments we use $\lambda=0$, i.~e., the loss given in Eq. \ref{eq:loss}, and therefore neglect joint limits.
\paragraph{Singularities} We evaluate effects of singularities on solutions to the IKP. We use a dataset which contains query poses that correspond to at least one singular configuration and another dataset from which near-singular configurations are removed entirely. Results are compared to a standard dataset which is uniformly sampled in the joint space and may contain both near-singular and nonsingular cases. It is well-known that numerical solvers built on the basis of the manipulator Jacobian tend to have problems near singular configurations as $\textrm{det}\,\bm{J} \approx 0$ and an inversion of $\bm{J}$ leads to very high joint velocities. For this reason it is interesting to see whether DT suffers from the same problem.
\paragraph{Approximation of unreachable poses} We investigate approximation of unreachable poses. Sometimes (e.~g., in reinforcement learning \cite{Fabisch2019}) it is beneficial if we can approximate poses outside of the workspace as close as possible. In the first part of this experiment, the training data lies in the workspace while the test data consists of end effector positions that are not reachable. In the second part, the test data stays the same, but some samples of the training data are swapped with poses that lie outside of the workspace.
\paragraph{Consistency} We analyze for DT and TRAC-IK how consistent the selection of the solution is. For this purpose we look at one trajectory in the Cartesian space. The analysis of discontinuities is performed in the joint space. With regard to practical applications it would be advantageous if discontinuities, i.~e., changes of the chosen IK solution for similar query poses, do not occur often. Note that in this investigation the output of the preceding query pose is always used as the initial configuration for TRAC-IK.
\paragraph{Runtime} We examine prediction runtime of DT and compare it with TRAC-IK and the analytical solution. 

\subsection{Evaluation Metrics}
We use the position error $\varepsilon_{pos}$ and the orientation error $\varepsilon_{ori}$ of the prediction (described in Section \ref{subsec:DTL}) as the metric for DT. Whenever the accuracy is compared to TRAC-IK, we calculate the solve rate for defined error thresholds. Regarding the investigation of how to handle joint limits in DT, we introduce the metric $\eta$ which describes the percentage of infeasible configurations from all predictions. A predicted configuration is considered infeasible if at least one joint angle violates the limits. In the experiment of approximation of unreachable poses the position and orientation error alone are not sufficient to judge how accurate the predictions are. As a comparative value, the mean euclidean distance from the base to the desired end effector position over all query poses, which is given by $\overline{\| \bm{x}^*_{pos} \|}$, is used. The vector $\bm{x}^*_{pos}$ consists of the desired Cartesian position coordinates of the end effector while the base of the mechanism is located at the origin of the coordinate system.

\section{Results and Discussion}
We tuned the network architecture per robot, resulting in dense hidden layers with 2x256 nodes (3-DoF arm), 6x512 nodes (COMPI), and 3x1024 nodes (Atlas).

\begin{figure}[t!]
\centering
\subfloat[Position errors.]{
  \includegraphics[width=0.33\textwidth]{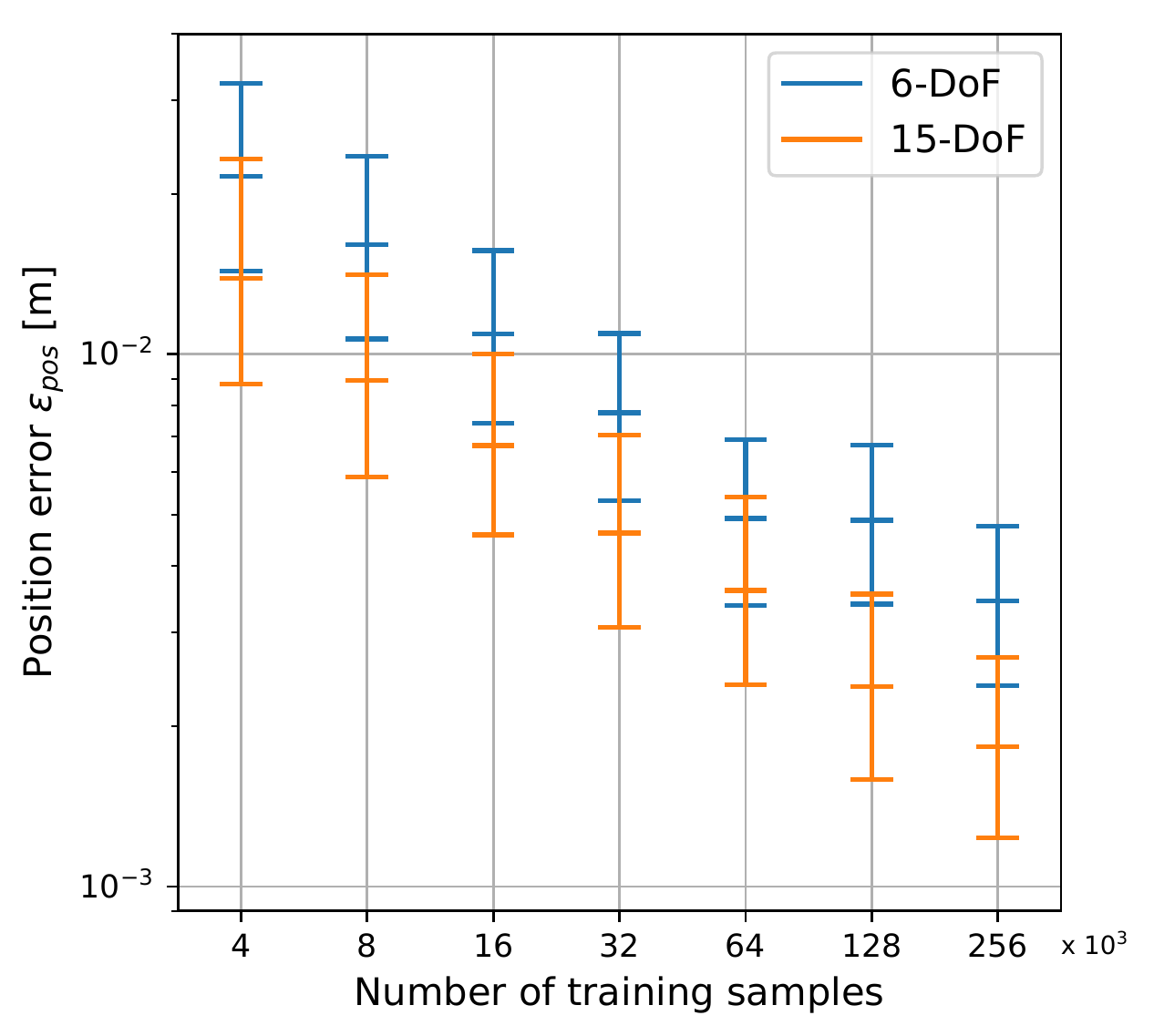}
	\label{fig:pos_errors_DT}
}\\
\vskip -0.02in
\subfloat[Orientation errors.]{
  \includegraphics[width=0.33\textwidth]{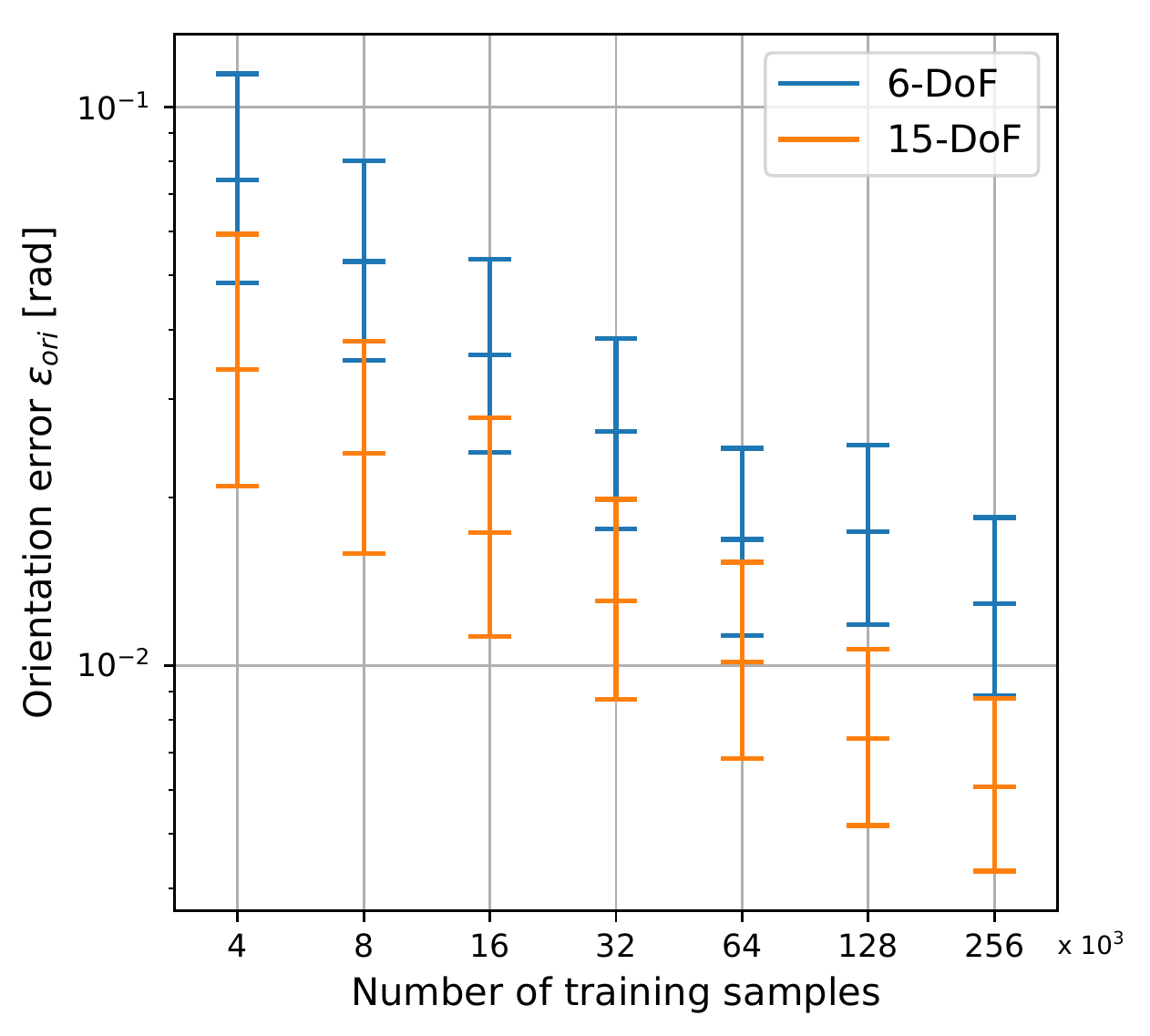}
	\label{fig:ori_errors_DT}
}
\caption{Dependency of Cartesian errors of DT on number of training samples for the 6-DoF and 15-DoF chains.\label{fig:cartesian_errors_DT}}
\vskip -0.25in
\end{figure}

\paragraph{Sample efficiency} Tables \ref{tab:accuracy_comparison} and \ref{tab:accuracy_comparison_low_threshold} show a comparison of the solve rates for different error thresholds and dataset sizes. The analytical solution is exact, but only available for the 3-DoF and 6-DoF manipulator. TRAC-IK finds less solutions the more complex the mechanism gets. If the error threshold is relatively large (cf. Table \ref{tab:accuracy_comparison}) and enough training data is used, DT can come close or even surpass (see 15-DoF) the numerical solve rate. However, in case of a low error threshold (cf. Table \ref{tab:accuracy_comparison_low_threshold}) DT performs far worse than TRAC-IK, even with several million training samples. Also note that DT -- in contrast to TRAC-IK -- finds more solutions for the redundant 15-DoF manipulator compared to the non-redundant 6-DoF mechanism. Fig. \ref{fig:cartesian_errors_DT} confirms this as it displays a direct comparison of the Cartesian errors between the 6-DoF and 15-DoF mechanism. It is further observable that the errors roughly get reduced by 50\% if the number of training samples is multiplied by four.

\begin{table}[t!]
\centering
\caption{
Evaluation of DT's sample efficiency. Tables list the percentage of the solutions below given error thresholds.
The number of training samples is shown in brackets.}
\vskip -0.05in
\subfloat[Error thresholds: \mbox{$\varepsilon_{pos} < 0.01 \, \textrm{m}$} and \mbox{$\varepsilon_{ori} < 0.03 \, \textrm{rad}$}.\label{tab:accuracy_comparison}]
{
\scalebox{0.95}{
\begin{tabular}{|l|l|r|r|r|r|}
\hline
Mechanism & Analy- & Numerical & DT    & DT    & DT \\
          & tical  & TRAC-IK & (100) & (800) & (6400)\\
\hline
3-DoF & 100\% & 100\% & 3.39\% & 62.88\% & 98.08\%\\
\hline
Mechanism & Analy- & Numerical & DT     & DT      & DT \\
          & tical  & TRAC-IK & (4000) & (32000) & (256000)\\
\hline
6-DoF  & 100\% & 98.44\% &  1.86\% & 53.58\% & 94.53\%\\
15-DoF & -     & 93.23\% & 18.13\% & 82.96\% & 97.73\%\\
\hline
\end{tabular}
}}\\
\vskip -0.05in
\subfloat[Error thresholds: \mbox{$\varepsilon_{pos} < 0.001 \, \textrm{m}$} and \mbox{$\varepsilon_{ori} < 0.01 \, \textrm{rad}$}.\label{tab:accuracy_comparison_low_threshold}]
{\scalebox{0.95}{
\begin{tabular}{|l|l|r|r|r|r|}
\hline
Mechanism & Analy- & Numerical & DT     & DT \\
          & tical  & TRAC-IK   & (1e+6) & (4e+6)\\
\hline
3-DoF & 100\% & 100\% & 76.49\% & 89.34\%\\
6-DoF  & 100\% & 98.44\% &  3.58\% & 9.59\%\\
15-DoF & -     & 93.23\% & 21.97\% & 59.65\%\\
\hline
\end{tabular}}}
\vskip -0.29in
\end{table}

\paragraph{Joint limits}
Results are listed in Table \ref{tab:joint_limits}. We can see that DT yields lots of infeasible predictions if we do not take joint limits into account ($\lambda=0$). The more joints are used, the higher the chance to violate at least one joint limit. If the penalty factor $\lambda$ is increased, the percentage of infeasible predictions $\eta$ decreases to about $1\%$ and for higher values of $\lambda$ potentially even further. The downside of this approach is a significant increase of the position and orientation errors. The highest increase is noticed for the 6-DoF mechanism where the errors increase approximately by a factor of four. This means that the sample efficiency of DT is worse if we consider joint limits as we need a much larger number of training samples to reach the same solve rates as depicted in Table \ref{tab:accuracy_comparison}. Nevertheless, with a known FK function we can generate a few million training samples quickly and using that many samples is not a problem with modern hardware.

\begin{table}[tb!]
\centering
\caption{Percentage of configurations with at least one infeasible joint angle $\eta$ and the mean and standard deviations of the position error $\varepsilon_{pos}$ and orientation error $\varepsilon_{ori}$ for different penalty factors $\lambda$ in DT. The number of training samples is 800 (3-DoF) or 32000 (6-DoF and 15-DoF).\label{tab:joint_limits}}
\scalebox{0.95}{
\begin{tabular}{|l|l|r|r|r|r|r|}
\hline
Mechanism & $\lambda$ & $\eta$ & $\varepsilon_{pos}$ [m] & $\varepsilon_{ori}$ [rad]\\
\hline
3-DoF & 0 & 12.81\% & 8.76e-3$\pm$8.25e-3 & 1.20e-2$\pm$1.41e-2\\
      & 1 & 1.27\% & 1.38e-2$\pm$2.67e-2 & 1.61e-2$\pm$3.09e-2\\
      & 2 & 0.97\% & 1.42e-2$\pm$2.60e-2 & 1.57e-2$\pm$2.91e-2\\
\hline
6-DoF & 0 & 71.22\% & 8.39e-3$\pm$6.80e-3 & 3.15e-2$\pm$5.03e-2\\
      & 1 & 3.30\% & 3.50e-2$\pm$4.57e-2 & 1.27e-1$\pm$1.53e-1\\
      & 2 & 1.89\% & 3.72e-2$\pm$5.28e-2 & 1.19e-1$\pm$1.55e-1\\
			& 5 & 1.16\% & 3.95e-2$\pm$5.47e-2 & 1.45e-1$\pm$2.22e-1\\
\hline
15-DoF & 0 & 100\% & 6.93e-3$\pm$9.67e-3 & 1.80e-2$\pm$3.46e-2\\
       & 1 & 6.12\% & 1.82e-2$\pm$3.80e-2 & 5.56e-2$\pm$1.47e-1\\
       & 2 & 3.02\% & 1.65e-2$\pm$3.01e-2 & 4.95e-2$\pm$1.20e-1\\
			 & 5 & 1.59\% & 1.31e-2$\pm$2.67e-2 & 4.13e-2$\pm$1.13e-1\\
\hline
\end{tabular}
}
\vskip -0.25in
\end{table}

\paragraph{Singularities} Results are shown in Table \ref{tab:singularities}. The analytical solution is exact. We expected that the numerical solver TRAC-IK is sensitive to singularities due to the pseudo-inverse of the Jacobian, which the resulting solve rates confirm. While there occur no failures for the simple planar 3-DoF mechanism, solve rates are clearly lower for singular configurations with the 6- and 15-DoF manipulators. On the dataset from which near-singular configurations were removed we also see an increase of the overall solve rate. DT seems to be more robust against singularities. While there are slight deviations in the solve rate, no clear trend can be seen so that we assume that these differences are not caused by singularities.

\begin{table}[tb!]
\centering
\caption{Solve rates for error thresholds \mbox{$\varepsilon_{pos} < 0.01 \, \textrm{m}$} and \mbox{$\varepsilon_{ori} < 0.03 \, \textrm{rad}$} and different test sets. Numbers in brackets below DT denote the number of training samples.\label{tab:singularities}}
\scalebox{0.95}{
\begin{tabular}{|l|l|l|r|r|r|}
\hline
Mecha-   & Test Set  & Analy- & Numerical & DT    & DT\\
nism     &           & tical  & TRAC-IK & (800) & (6400)\\
\hline
3-DoF & Uniform & 100\% & 100\% & 62.88\% & 98.08\%\\
      & Singular & 100\% & 100\% & 61.81\% & 99.08\%\\
      & Nonsingular & 100\% & 100\% & 59.17\% & 97.25\%\\
\hline
Mecha-   & Test Set   & Analy- & Numerical & DT      & DT\\
nism     &            & tical  & TRAC-IK & (32000) & (256000)\\
\hline
6-DoF & Uniform & 100\% & 98.44\% & 53.58\% & 94.53\%\\
      & Singular & 100\% & 86.95\% & 52.27\% & 95.26\%\\
      & Nonsingular & 100\% & 99.08\% & 51.58\% & 95.98\%\\
\hline
15-DoF & Uniform & - & 93.23\% & 82.96\% & 97.73\%\\
       & Near-sing. & - & 89.96\% & 78.21\% & 96.60\%\\
       & Nonsingular & - & 95.34\% & 82.96\% & 98.12\%\\
\hline
\end{tabular}
}
\vskip -0.1in
\end{table}

\paragraph{Approximation of unreachable poses}
Results are listed in Table \ref{tab:unreachable_poses}. If we do not use similar unreachable poses in the training data, the predictions are of low quality. The mean position error is a bit lower than the comparative value $\overline{\| \bm{x}^*_{pos} \|}$ which means that -- on average -- the arm is stretching out into the right direction, but the difference of $\overline{\| \bm{x}^*_{pos} \|}$ and $\varepsilon_{pos}$ is nowhere near the total arm length. Also the orientation errors are very high. As we swap some of the training samples with unreachable poses in a similar range of the query poses, all of the errors are decreasing clearly. Now the difference of $\overline{\| \bm{x}^*_{pos} \|}$ and $\varepsilon_{pos}$ gets close to the total arm length for each mechanism (be aware that the difference cannot reach the total arm length because the last link of the manipulator needs to be used to approximate the desired orientation). Furthermore, sharp declines of the orientation errors are noticeable but these errors remain high compared to the errors for query poses in the workspace (cf. Table \ref{tab:joint_limits}). Consequently, an approximation of unreachable poses with DT is possible but to reach very accurate predictions a large proportion of the training data needs to lie outside of the workspace. While not directly investigated, we can assume that this impairs the predictions for query poses inside of the workspace. Note that the applied numerical solution is not directly suitable to solve for poses outside of the workspace but it can be modified to do this accurately (e.g., \cite{Fabisch2019}).

\begin{table}[tb!]
\centering
\caption{Means and standard deviations for the position error $\varepsilon_{pos}$ and orientation error $\varepsilon_{ori}$ for unreachable query poses without and partly with unreachable poses (u.p.) in the training data of DT. The number of training samples is 800 (3-DoF) or 32000 (6-DoF and 15-DoF).\label{tab:unreachable_poses}}
\begin{tabular}{|l|l|l|l|l|}
\hline
Mechanism & Training & $\overline{\| \bm{x}^*_{pos} \|}$ [m] & $\varepsilon_{pos}$ [m] & $\varepsilon_{ori}$ [rad]\\
          & Dataset  &                                       &                         &                          \\
\hline
3-DoF & No u.p.  & 3.25 & 2.54$\pm$0.47 & 0.95$\pm$0.76\\
      & 100 u.p. & 3.25 & 2.25$\pm$0.43 & 0.10$\pm$0.09\\
\hline
6-DoF & No u.p.   & 1.95 & 1.62$\pm$0.35 & 1.95$\pm$0.70\\
      & 1000 u.p. & 1.95 & 1.17$\pm$0.25 & 0.66$\pm$0.46\\
\hline
15-DoF & No u.p.   & 5.28 & 4.51$\pm$0.73 & 2.27$\pm$0.62\\
       & 1000 u.p. & 5.28 & 3.27$\pm$0.62 & 0.44$\pm$0.43\\
\hline
\end{tabular}
\vskip -0.25in
\end{table}

\begin{figure}[tb!]
\vskip -0.1in
\centering
\subfloat[6-DoF, distal teaching.]{
  \includegraphics[width=0.22\textwidth]{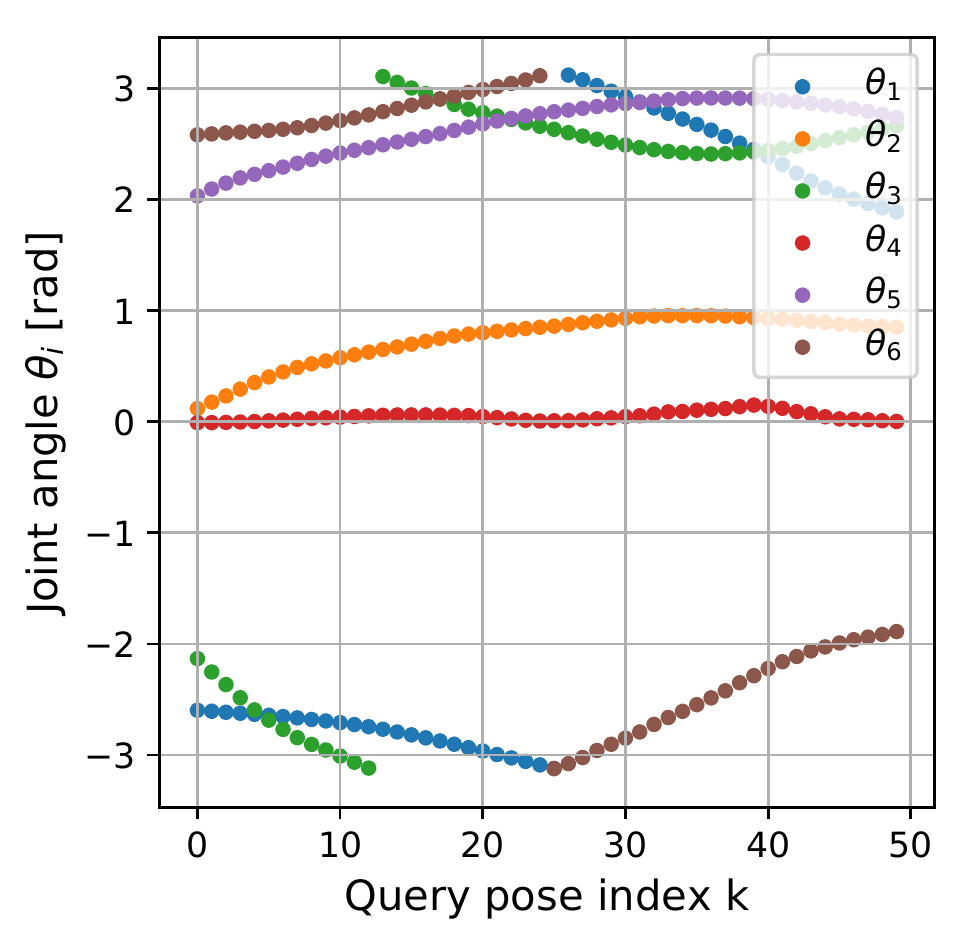}
	\label{fig:trajectory_DT_6DOF}
}
\subfloat[6-DoF, TRAC-IK.]{
  \includegraphics[width=0.22\textwidth]{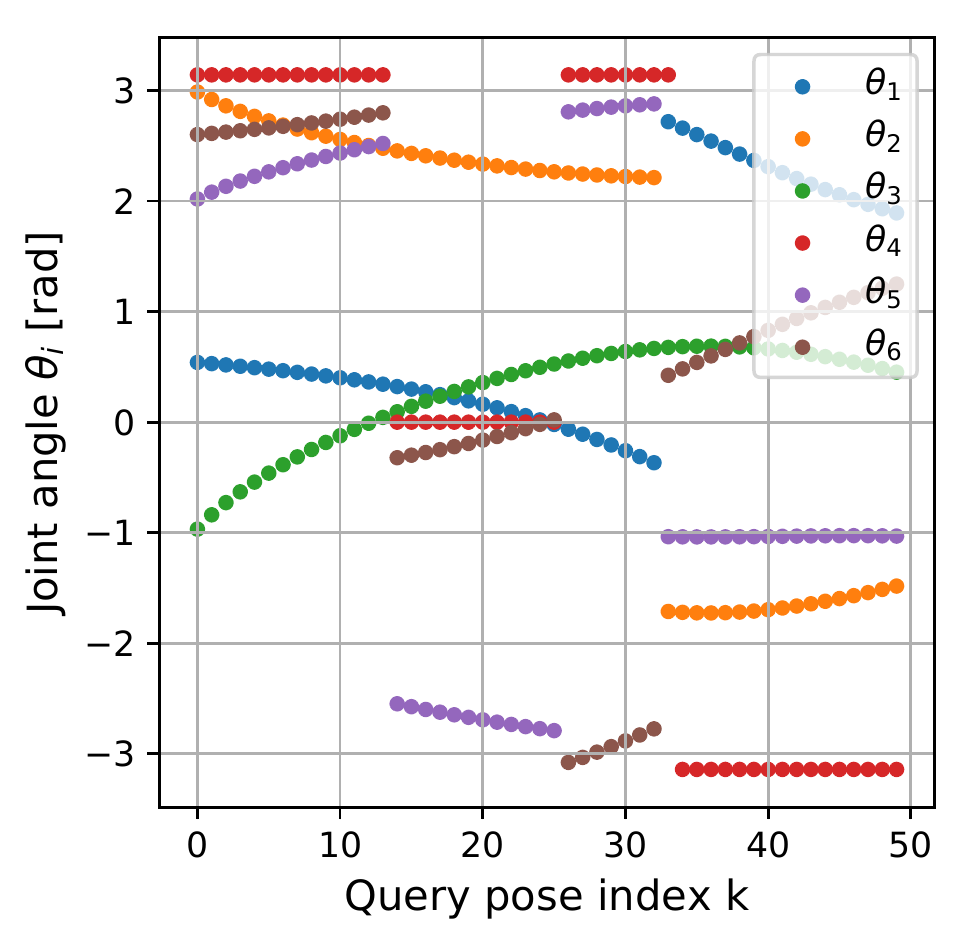}
	\label{fig:trajectory_TRACIK_6DOF}
}\hfil
\vskip -0.05in
\subfloat[15-DoF, distal teaching.]{
  \includegraphics[width=0.22\textwidth]{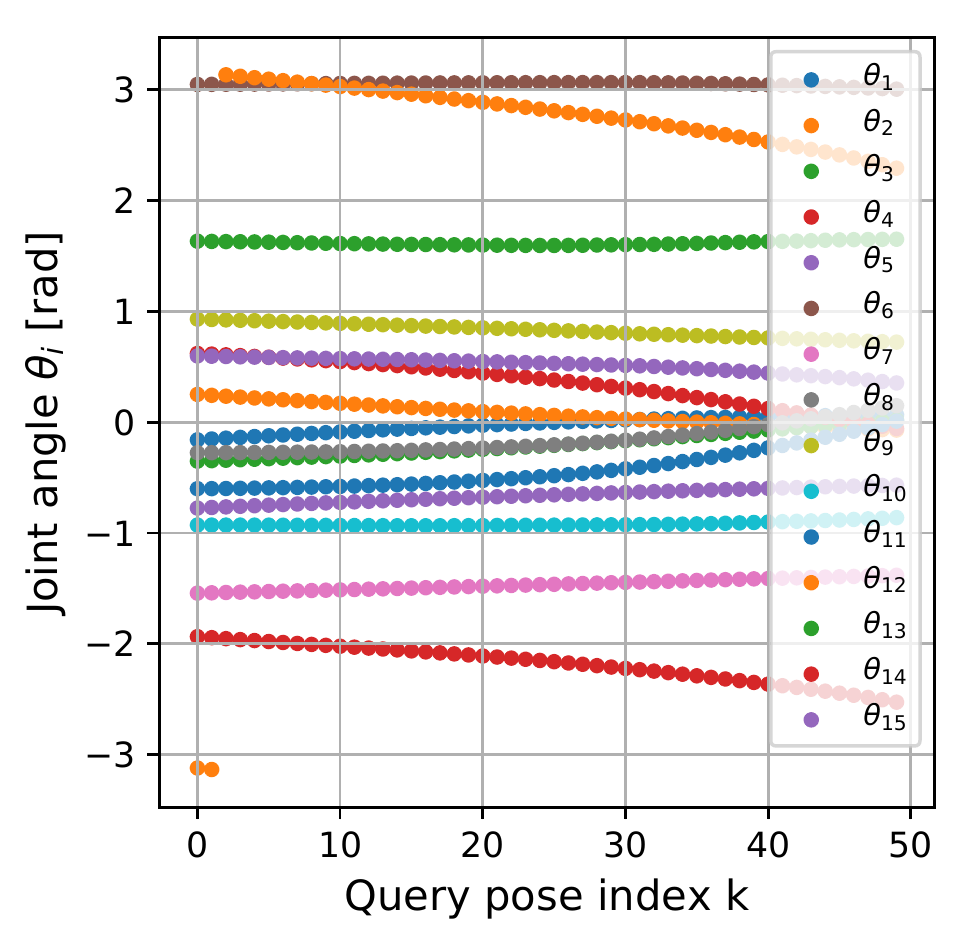}
	\label{fig:trajectory_DT_15DOF}
}
\subfloat[15-DoF, TRAC-IK.]{
  \includegraphics[width=0.22\textwidth]{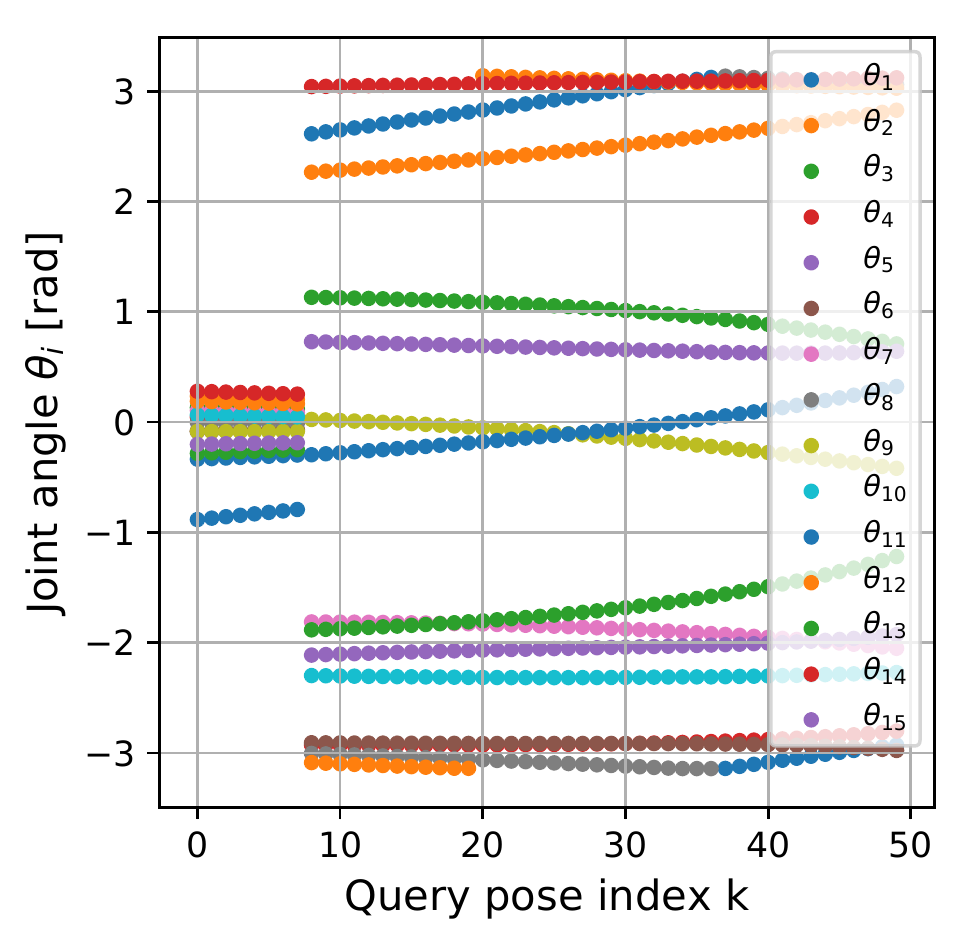}
	\label{fig:trajectory_TRACIK_15DOF}
}
\caption{Solutions to the IKP in joint space for one trajectory.\label{fig:solution_choice}}
\vskip -0.25in
\end{figure}

\paragraph{Consistency}
Fig. \ref{fig:solution_choice} compares the predictions in joint space for one end effector trajectory per mechanism. For the 3-DoF arm (not shown) both methods achieve an optimal continuity. In the 6-DoF-case DT remains continuous but TRAC-IK shows a few discontinuities, i.e., changes of the chosen IK solution. Regarding the 15-DoF chain both results are relatively good. No discontinuity occurs when DT is used and only one if TRAC-IK is applied. While these cases are only examples, it seems like DT does not switch the IK solution as often as TRAC-IK. This is an advantageous property of DT but note that TRAC-IK cannot be seen as a representative of all numerical solvers because one of TRAC-IK's main objectives is to maximize the solve rate. Its SQP-solver does not use information about the preceding joint configuration \cite{Beeson.2015}.

\paragraph{Runtime}
In Table \ref{tab:runtime_comparison}, runtime for sequential processing of query poses with DT is compared with the other approaches. It is very clear that the analytical solution is by far the fastest and most constant method -- if it is available. The numerical solver TRAC-IK needs much longer in general. We can observe an increase of the runtime with the complexity of the manipulator. The runtime also fluctuates a lot as there is a strong dependence on the specific query pose (cf. the singular data set) while the runtimes of the analytical solution and DT do not depend on the input. The latter can be considered as quite slow for simple mechanisms though. In DT the runtime strongly depends on the neural network architectures which we tuned manually. As the architecture was optimized for each manipulator individually, the runtimes differ. In the 15-DoF case, DT has the overall best runtime.
Due to the fact that neural networks can often be calculated more efficiently on GPUs in parallel, we investigated some more cases for DT (Table \ref{tab:runtime_DT}). It turns out that the usage of a GPU results in faster runtimes in case of the 6-DoF and 15-DoF mechanism for which rather large neural networks are used. Furthermore, processing query poses as batches of 32 nearly reduces the runtime by a factor of the batch size. This can be beneficial if IK needs to be solved for full trajectories at once.

Another important aspect to be considered in a comparison to traditional solutions is the training time of DT. For the 15-DoF chain and four million training samples, the training time on a modern GPU is roughly two weeks. 

\begin{table}[tb!]
\centering
\caption{Means and standard deviations of the elapsed real time to solve the IKP.
CPU: Intel Core i7-6800K; GPU: Nvidia Gefore GTX 1080 Ti.}
\vskip -0.05in
\subfloat[Computation times for sequential processing on a CPU.\label{tab:runtime_comparison}]
{
\scalebox{0.95}{
\begin{tabular}{|l|l|l|l|l|}
\hline
Method     & Test Set  & 3-DoF [ms]      & 6-DoF [ms]      & 15-DoF [ms]\\
\hline
Analytical & Uniform   & 0.009$\pm$0.001 & 0.053$\pm$0.002 & - \\
\hline
Numerical  & Uniform   & 0.179$\pm$0.085 & 0.908$\pm$0.916 & 1.717$\pm$1.353\\
TRAC-IK    & Singular  & 0.193$\pm$0.055 & 1.533$\pm$1.664 & 2.019$\pm$1.449\\
           & Nonsing.  & 0.181$\pm$0.096 & 0.903$\pm$0.868 & 1.582$\pm$1.292\\
\hline
DT         & Uniform   & 0.523$\pm$0.228 & 1.774$\pm$0.310 & 1.610$\pm$0.308\\
\hline
\end{tabular}
}
}\\
\vskip -0.05in
\subfloat[Computation times for DT using different ways to process queries.\label{tab:runtime_DT}]
{
\scalebox{0.95}{
\begin{tabular}{|l|l|l|l|l|}
\hline
Processing & Device & 3-DoF [ms]      & 6-DoF [ms]      & 15-DoF [ms]\\
\hline
Sequential    & CPU & 0.523$\pm$0.228 & 1.774$\pm$0.310 & 1.610$\pm$0.308\\
Batch (32)    & CPU & 0.014           & 0.054           & 0.061\\
Sequential    & GPU & 0.983$\pm$0.239 & 1.520$\pm$0.099 & 1.180$\pm$0.113\\
Batch (32)    & GPU & 0.017           & 0.046           & 0.039\\
\hline
\end{tabular}
}
}
\vskip -0.35in
\end{table}

\section{Conclusions}
We analyzed distal teaching (DT), a data-driven solution to the IKP, in case of rigid-body mechanisms. For this purpose, we propose the integration of the usually known forward kinematics into the loss function of a feed-forward neural network. Further, we compared the results to an analytical solution as well as TRAC-IK, a numerical solver.

The results show that analytical solutions are superior due to their exactness, low and constant runtime as well as the provision of all solutions. However, they are unavailable for general redundant manipulators. If high accuracy matters or the degree of redundancy is fairly low, we reason that numerical solutions are the best choice. For rather simple mechanisms, they have shown to be faster and to achieve better solve rates than DT, especially for high accuracy requirements. DT only makes sense in case of rigid-body mechanisms if those are highly redundant and accuracy requirements are relaxed. With the example of the 15-DoF serial chain of Atlas, we have shown that reaching higher solve rates than TRAC-IK is possible while offering lower and less fluctuating runtimes. Such a performance is only achievable with a large amount of training data ($>10^6$) though. While these are easy to generate under known forward kinematics, this goes along with weeks of training. Dealing with joint limits and accurate approximation of unreachable poses increases the need of data even further. Positive properties of DT are the robustness against singularities and the continuous choice of solutions.
Numerical IK solvers are better in terms of reconfigurability than analytical solutions or learned IK solutions
as they can deal with generic robot geometry while analytical solutions may not be valid if there is a significant change in robot design and machine learning based IK solutions would need retraining.

As a side note, there is an interesting similarity between numerical solvers and machine learning (ML): while numerical solvers find one solution to an optimization problem online, ML solves the same optimization problem offline and stores a set of solutions, in our case, in a neural network.

In the future, we will need complex robotic systems that are hard to model and control accurately (e.~g., soft robots \cite{Yu2019}, series-parallel hybrid robots with many DoF \cite{Bartsch2016,2019_recupera_full_exo}) to act in highly dynamic environments, which poses a problem to both purely classical and ML based approaches. However, we believe that combining classical and ML-based solutions in an optimal way to realize what should be called hybrid artificial intelligence is the way forward.

\addtolength{\textheight}{-12cm}  % This command serves to balance the column lengths

%%%%%%%%%%%%%%%%%%%%%%%%%%%%%%%%%%%%%%%%%%%%%%%%%%%%%%%%%%%%%%%%%%%%%%%%%%%%%%%%

%%%%%%%%%%%%%%%%%%%%%%%%%%%%%%%%%%%%%%%%%%%%%%%%%%%%%%%%%%%%%%%%%%%%%%%%%%%%%%%%

%%%%%%%%%%%%%%%%%%%%%%%%%%%%%%%%%%%%%%%%%%%%%%%%%%%%%%%%%%%%%%%%%%%%%%%%%%%%%%%%
%\section*{APPENDIX}

\section*{ACKNOWLEDGMENT}
We thank Matias Valdenegro-Toro for helpful feedback.

%The third author acknowledges the support of Q-RoCK project (Grant number: FKZ 01IW18003) funded by the German Aerospace Center (DLR) with federal funds from the Federal Ministry of Education and Research (BMBF).

%%%%%%%%%%%%%%%%%%%%%%%%%%%%%%%%%%%%%%%%%%%%%%%%%%%%%%%%%%%%%%%%%%%%%%%%%%%%%%%%

%\begin{thebibliography}{99}
%\bibitem{c1} Reference 1
%\end{thebibliography}

\bibliographystyle{./IEEEtran} % use IEEEtran.bst style
\bibliography{./IEEEabrv,./IEEEexample}

\end{document}